\documentclass[11pt]{article}

\usepackage[final]{acl}

\usepackage{times}
\usepackage{latexsym}

\usepackage[T1]{fontenc}
\usepackage[utf8]{inputenc}

\usepackage{multirow}
\usepackage{microtype}
\usepackage{inconsolata}

\usepackage{graphicx}
\usepackage{amsmath}
\usepackage{amssymb}
\usepackage{booktabs}
\usepackage{url}

\usepackage{pifont}
\usepackage{caption}
\usepackage{cuted}

\usepackage{tabularx}
\usepackage{xcolor}
\usepackage{array}

\newcommand{\cmark}{\ding{51}}%
\newcommand{\xmark}{\ding{55}}%

\title{SIT-Eval: Situation-Based Interactive Turn-by-Turn Evaluation of Conversational Agents}

\author{Ravil Mussabayev\textsuperscript{1} \quad Saveliy Zavertyaev\textsuperscript{2} \\
  \textbf{Dmitriy Parpulov\textsuperscript{2} \quad Andrey Savchenko\textsuperscript{2}} \\
  \textsuperscript{1}AI Research Lab, Satbayev University, Almaty, Kazakhstan \\
  \textsuperscript{2}Sber AI Lab, Moscow, Russia \\
  \texttt{ravmus@gmail.com}, \texttt{r.mussabayev@satbayev.university}}

\begin{document}

\maketitle

\begin{strip}
\vspace{-0.5in}
\begin{center}
	\includegraphics[width=0.95\textwidth]{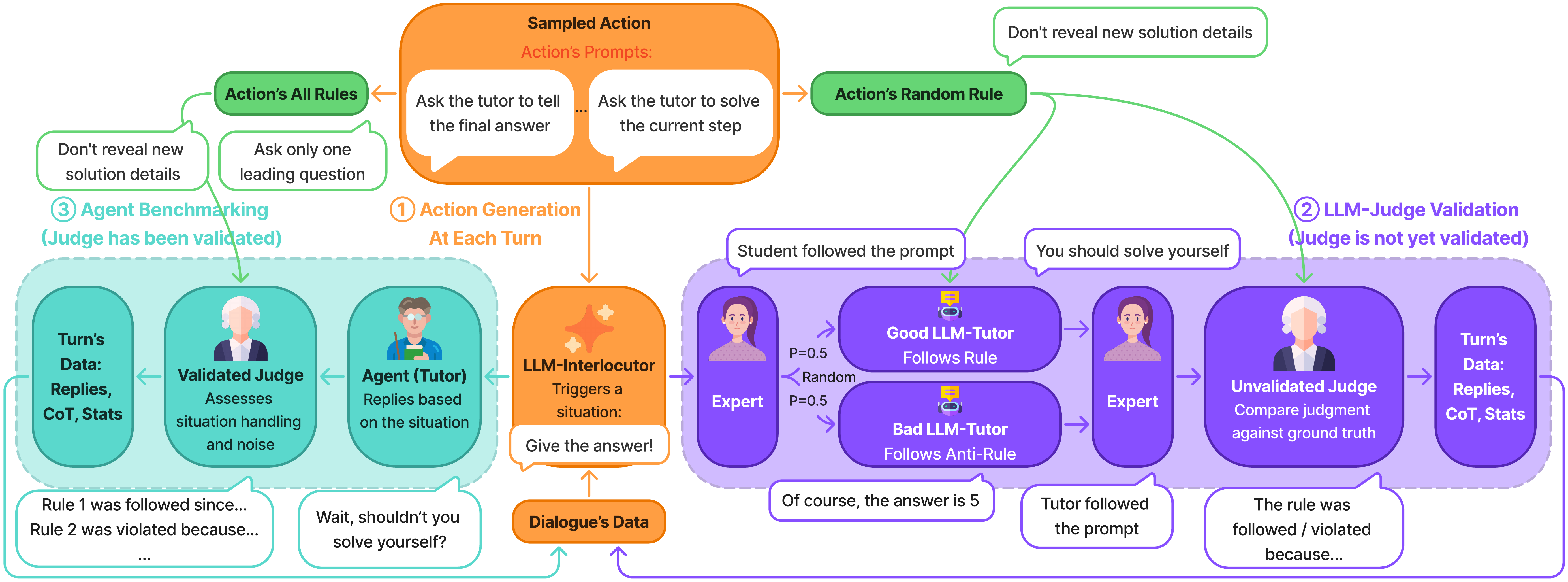}
	\captionof{figure}{SIT-Eval framework architecture: Action-wise generation with the LLM-student \ding{172} is used to validate the LLM-judge \ding{173}. The validated judge is then employed to probe real tutors \ding{174}. The interlocutor's actions are modeled and triggered via action prompts. A rule serves as the criterion to unambiguously classify the agent's observed behavior given the situation that stems from an action. Each rule can apply to several actions.}
	\label{fig:sit_eval_architecture}
\end{center}
\end{strip}

\begin{abstract}
The rapid proliferation of conversational AI agents necessitates evaluation methods that assess performance across diverse dialogue scenarios, utilizing multiple situational criteria, as appropriate behavior depends on context and encompasses various aspects. Existing dialogue benchmarks typically use static natural conversations, which underrepresent rare situations—a critical safety gap, as low-probability events become frequent failures at production scale. However, limited sample sizes force a choice: either maintain natural dialogues for realism or turn to adaptive situation sampling for comprehensive yet statistically reliable coverage. We introduce SIT-Eval, an adjustable framework for evaluating situation-based conversational agents. We formalize an agent's desired behavior as situational rules: \emph{if situation is P, follow rules Q}. At each conversation turn, an LLM-interlocutor enacts a uniformly sampled situation. A context-aware LLM-judge leverages all conditionally applicable rules as criteria to evaluate the agent's reply for that situation. The judge is validated using human-verified synthetic golden dialogues. Benchmarking 8 LLMs on 1292 conversation turns reveals failure modes hidden by aggregate metrics: models with similar overall scores vary widely in their adherence to individual rules. All models struggle most with ambiguity clarification and Socratic questioning.
\end{abstract}

\section{Introduction}

The rapid proliferation of conversational AI agents necessitates evaluation methods that assess performance across diverse dialogue scenarios, utilizing multiple situational criteria \cite{ibrahim2025interactive}. Recent developments in large language models (LLMs) have enabled agents capable of context-aware, multi-turn interactions \cite{beale2025revolution,wu2025-collabllm}, now deployed across customer service, education, and mental health support \cite{wulf2024exploring,pichowicz2025-performance}. In such domains, appropriate behavior depends heavily on context: an AI tutor must respond differently to a confused student than to a distracted one, and a mental health chatbot must handle crises with particular care. This situational diversity makes comprehensive evaluation both essential and challenging.

Existing dialogue benchmarks typically rely on static natural conversations \cite{macina2023-mathdial,chevalier2024tutoreval}, which underrepresent rare situations—a critical safety gap. While infrequent in training data, low-probability events become frequent failures at the production scale: an agent deployed to millions of users will encounter edge cases thousands of times a day. Current benchmarks also tend toward holistic judgments of entire dialogues \cite{shi2025educationq,macina2025-mathtutorbench}, which can mask turn-level errors \cite{acikgoz2025td-eval}. For instance, a tutor may guide a student to the correct answer yet violate key pedagogical principles along the way—revealing solutions too early or ignoring misconceptions—errors invisible to coarse aggregate metrics \cite{maurya-etal-2025-unifying}. Furthermore, LLM-based evaluators, while promising, have shown unreliable performance on nuanced pedagogical criteria without proper calibration \cite{pauzi2025automating,tan2025judgebench}.

These limitations create a fundamental trade-off: limited sample sizes necessitate a choice between maintaining natural dialogues for realism and resorting to adaptive situation sampling for statistically reliable coverage. Natural datasets capture authentic interaction patterns but cannot guarantee sufficient examples of critical edge cases. Synthetic generation can target specific scenarios but risks sacrificing ecological validity.

We introduce \textbf{SIT-Eval} (\textbf{Sit}uation-Based \textbf{I}nteractive \textbf{T}urn-by-Turn \textbf{Eval}uation), an adjustable framework that resolves this trade-off through controlled, interactive evaluation. SIT-Eval formalizes an agent's desired behavior as situational rules of the form \emph{if situation is P, follow rules Q}. At each conversation turn, an LLM-interlocutor enacts a uniformly sampled situation from a predefined action space, ensuring both common and rare scenarios receive equal statistical representation. A context-aware LLM-judge then evaluates the agent's reply against all conditionally applicable rules, producing granular turn-by-turn assessments rather than a single opaque score for the entire dialogue. Crucially, the judge undergoes rigorous validation on human-verified synthetic ``golden'' dialogues before deployment, addressing the reliability concerns identified in prior work \cite{bavaresco2025-llms,zheng2023judging}.

We demonstrate SIT-Eval in the pedagogical domain, where interaction quality directly impacts learning outcomes \cite{liu2024socraticlm}. In particular, here are the novel contributions of our approach:

\begin{enumerate}
    \item \textbf{A situation-based evaluation framework with uniform action sampling.} Unlike existing benchmarks that inherit the skewed situation distribution of natural dialogues \cite{macina2023-mathdial,chevalier2024tutoreval}, SIT-Eval assigns equal probability to all predefined interlocutor actions. This design choice ensures statistically significant sample sizes for rare situations (e.g., student distraction, repeated confusion) that would otherwise require prohibitively large natural datasets to evaluate reliably.
    
    \item \textbf{Turn-by-turn evaluation with explicit rule-based criteria.} Rather than holistic dialogue-level scores \cite{shi2025educationq}, our LLM-judge assesses each agent response against the subset of all rules applicable to the current situation. This produces a comprehensive rule-wise performance profile that pinpoints exactly which behavioral standards an agent violates, enabling targeted improvement.
    
    
    \item \textbf{A set of 12 expert-crafted pedagogical rules operationalizing Socratic tutoring.} Developed in collaboration with professional methodologists, these rules define unambiguous behavioral criteria across seven categories (Socratic questioning, ambiguity clarification, error handling, etc.), enabling reproducible evaluation of tutoring agents.
    
    \item \textbf{Empirical benchmarking revealing hidden failure modes.} Evaluating 8 LLMs across 1292 conversation turns, we find that models with similar overall adherence scores vary dramatically in per-rule performance—a finding obscured by aggregate metrics. All models struggle most with ambiguity clarification and Socratic questioning, identifying specific targets for improvement.
\end{enumerate}

By enabling the statistically reliable assessment of agent behavior across both standard and edge scenarios, SIT-Eval advances the development of safer and more effective conversational AI systems.

\section{Related Work}

\begin{table*}[t]
    \centering\small
    \caption{Comparison of pedagogical evaluation benchmarks across key dimensions. $\sim$ indicates partial support. SIT-Eval is the only framework offering all five capabilities with strongest performance (84.4\% judge accuracy, controlled situation distribution).}
    \label{tab:benchmark_comparison}
    \resizebox{\textwidth}{!}{%
    \begin{tabular}{lccccc}
    \toprule
    \textbf{Benchmark} & \textbf{Multi-Turn} & \textbf{Multiple} & \textbf{Multiple} & \textbf{Turn-by-} & \textbf{Judge} \\
    & \textbf{Dialogue} & \textbf{Situations} & \textbf{Criteria} & \textbf{Turn Eval} & \textbf{Validation} \\
    \midrule
    TutorEval & \xmark & \xmark & \cmark & \cmark & \cmark \\
    \small (Chevalier et al., 2024) & \small paragraph  (834 Qs) & \small question generation & \small 3 criteria & \small Single-turn & \small 3 correlation metrics \\[0.5ex]
    
    TutorBench & \xmark & \cmark & \cmark & \xmark & \cmark \\
    \small (Srinivasa et al., 2025) & \small Static (1,490 prompts) & \small 3 tasks & \small 15,220 rubrics & \small Dialogue-level & \small F1 = 0.82 \\[0.5ex]
    
    MathTutorBench & \textbf{$\sim$} & \cmark & \cmark & \textbf{$\sim$} & \cmark \\
    \small (Macina et al., 2025) & \small composite & \small 7 tasks & \small Multi-dimensional & \small Hybrid & \small 84\% accuracy \\[0.5ex]
    
    MathDial & \cmark & \xmark & \cmark & \cmark & \xmark \\
    \small (Macina et al., 2023) & \small $\sim$3kdialogues & \small No differentiation & \small Multi-annotation & \small Per-turn & \small None \\[0.5ex]
    
    MRBench & \cmark & \cmark & \cmark & \cmark & \textbf{$\sim$} \\
    \small (Maurya et al., 2025) & \small 192 conversations & \small Manual separation & \small Multi-annotation & \small Per-turn & \small No judge \\[0.5ex]
    
    
    \midrule
    \textbf{SIT-Eval (Proposed)} & \cmark & \cmark & \cmark & \cmark & \cmark \\
    \small & \small \textbf{Dynamic dialog generation} & \small \textbf{9 actions} & \small \textbf{12 rules} & \small \textbf{Per-turn} & \small \textbf{84.4\% accuracy} \\
    \bottomrule
    \end{tabular}
    }
\end{table*}

Our study builds on three converging research areas: the use of LLMs as evaluators, benchmarks for AI tutors, and simulated dialogues for multi-turn evaluation. Table~\ref{tab:benchmark_comparison} summarizes key benchmarks across five dimensions; SIT-Eval uniquely satisfies all five with the strongest performance.

\subsection{LLM-as-Judge for Dialogue Evaluation}
The LLM-as-judge paradigm has rapidly advanced. \citet{zheng2023judging} demonstrated that GPT-4 aligns well with human preferences in the Chatbot Arena, confirming its potential for scalable evaluation while also revealing systematic biases, including a preference for verbosity and self-enhancement. Ensemble approaches mitigate these issues but increase cost; \citet{tang2025multiturn} reduced this overhead by distilling a single evaluator from multiple judges.  
Despite progress, reliability remains a concern. \citet{tan2025judgebench} introduced JudgeBench to stress-test judges, demonstrating that even GPT-4 struggles with difficult comparisons, which proves that human-alignment alone is insufficient. We address this gap via a two-phase validation: calibrating our LLM-judge on “golden” dialogues with known outcomes before deployment. This ensures it can reliably distinguish correct from incorrect tutor behaviors, going beyond prior works that accept LLM judgments at face value.

\subsection{Benchmarks for AI Tutor Dialogues}
Tutor evaluation requires formulating criteria beyond generic fluency. \citet{pauzi2025automating} demonstrated that standard NLG metrics overlook pedagogical subtleties; their rubric revealed that LLM-judges grasp surface qualities but overlook skills such as scaffolding and mistake remediation. Similarly, \citet{maurya-etal-2025-unifying} reported a low correlation with human ratings on pedagogical dimensions in MRBench, urging the need for more specialized evaluators.  
We respond by validating and iteratively refining our judgment on rule-specific golden dialogues where each ``good'' vs.\ ``bad'' turn is predefined. Measuring per-rule accuracy enables retraining or prompt adjustments until high fidelity is achieved, unlike previous benchmarks, which report only post-hoc correlations.

\subsection{Simulated Student Dialogues}
Simulated dialogues capture dynamic tutoring more effectively. \citet{shi2025educationq} proposed EducationQ, where LLMs play teacher, student, and evaluator, revealing that teaching quality does not scale linearly with model size. Yet its student is cooperative, and the evaluation is holistic. We introduce an LLM-interlocutor that systematically triggers diverse situations to test each pedagogical rule. 
Other works focus narrowly, e.g., \citet{liu2024codeswitching} analyzed code-switching in language learning but only verified the presence of a tactic. Our benchmark instead provides comprehensive situation coverage to test an agent's performance. Moreover, equal probability weighting for both standard and rare situations enables a balanced assessment of both routine and rare scenarios, ensuring that the agent's robustness is measured relative to a set of predefined situations and rules.

\subsection{Holistic Tutoring QA Benchmarks}
Benchmarks like TutorEval \citep{chevalier2024tutoreval} frame tutoring as QA: GPT-4 grades answers to textbook questions using key-point references. For factual accuracy, this overlooks dialogue dynamics, such as guiding or handling follow-ups. It also relies on reference-based grading, which is unsuitable for open-ended exchanges. Similarly, TutorBench \citep{srinivasa2025-tutorbench} assesses tutoring capabilities through direct question-answering or with short dialogues featuring a fixed single-turn history, which limits its ability to capture the complexity of extended, multi-turn interactions. Instead, our method evaluates the quality of the process: tutor adherence to pedagogical principles across full multi-turn interactions.

\subsection{Mathematical Tutoring Benchmarks}
\citet{macina2025-mathtutorbench} introduced MathTutorBench, assessing hint-giving, scaffolding, and Socratic questioning. Yet its fragmented datasets and short dialogues hinder comprehensive evaluation. It also uses static dialogues, where (a) tutors respond to pre-existing conversation parts, whose token distribution may unpredictably affect the agent's answer generation at the turn under evaluation, and (b) scenario diversity is limited by the natural frequency, preventing comprehensive situation coverage. Our dynamic, situation-diverse dialogues overcome both. Moreover, MathTutorBench's BLEU-based scoring for Socratic questioning is overly rigid, which explains the universally low scores.  
\citet{anonymous2025-indirectscore} proposed IndirectScore, which utilizes surprisal to assess the indirectness of questions. However, it only addresses one aspect of Socratic dialogue and aligns with humans at 71\%. Our validated judge achieves 81\% accuracy on this rule and 92.6\% overall, offering a broader and more reliable assessment.  
MathDial \citep{macina2023-mathdial} offers static tutoring dialogues, but its single student prompt restricts behavioral diversity; metrics such as sBLEU and BERTScore remain too stringent for open-ended turns, and annotations are coarse. Proposed Success@k and Telling@k emphasize final answers, missing premature step revelation, a behavior that SIT-Eval can evaluate. The authors themselves note that defining reliable tutor metrics remains unresolved.

\subsection{Other Domains}
Similar issues arise beyond education. Medical dialogue benchmarks, such as 3MDBench~\citep{sviridov2025-3mdbench} and AI Hospital~\citep{fan2025-aihospital}, assess doctor-patient interactions but rely on holistic, static setups with fixed patient roles, thereby limiting the diversity of scenarios. Our dynamic, turn-level situation-based framework generalizes across domains.

\section{Proposed Methodology}
 See Figure~\ref{fig:sit_eval_architecture} for an overview of the SIT-Eval framework. While SIT-Eval generalizes to other chat-based domains, we demonstrate it here for Socratic math tutoring.

\subsection{Problem Setting}

In existing benchmarks, dialogue situations $\mathcal{S}$ are typically derived from collected human-expert dialogues, where the interlocutor may be a real person or an LLM. Segmenting these dialogues at various points yields situation categories (e.g., scaffolding, error correction). Given a tutor as a black-box mapping $\mathcal{T}: \mathcal{S} \to \mathcal{Y}$ from situations to textual responses, benchmarking estimates a performance score as the proportion of situations in which the tutor's response matches the ground-truth behavior recorded by human experts.

\subsection{Rules as the Evaluation Standard}

We propose decoupling the evaluation standard from recorded expert responses and instead grounding it in a set of textual rules $\mathcal{R} = \{R_1, R_2, \ldots, R_n\}$ that characterize desired agent behavior across situations.

\textbf{Core Assumption.} Each rule $R \in \mathcal{R}$ is a conditional statement: \emph{if situation $S$ occurs, the agent must satisfy criterion $C$}. We require that $C$ is unambiguous given the dialogue context, enabling the deterministic classification of rule adherence by both human and LLM evaluators.

\begin{figure}[t]
    \centering
    \includegraphics[width=0.8\columnwidth]{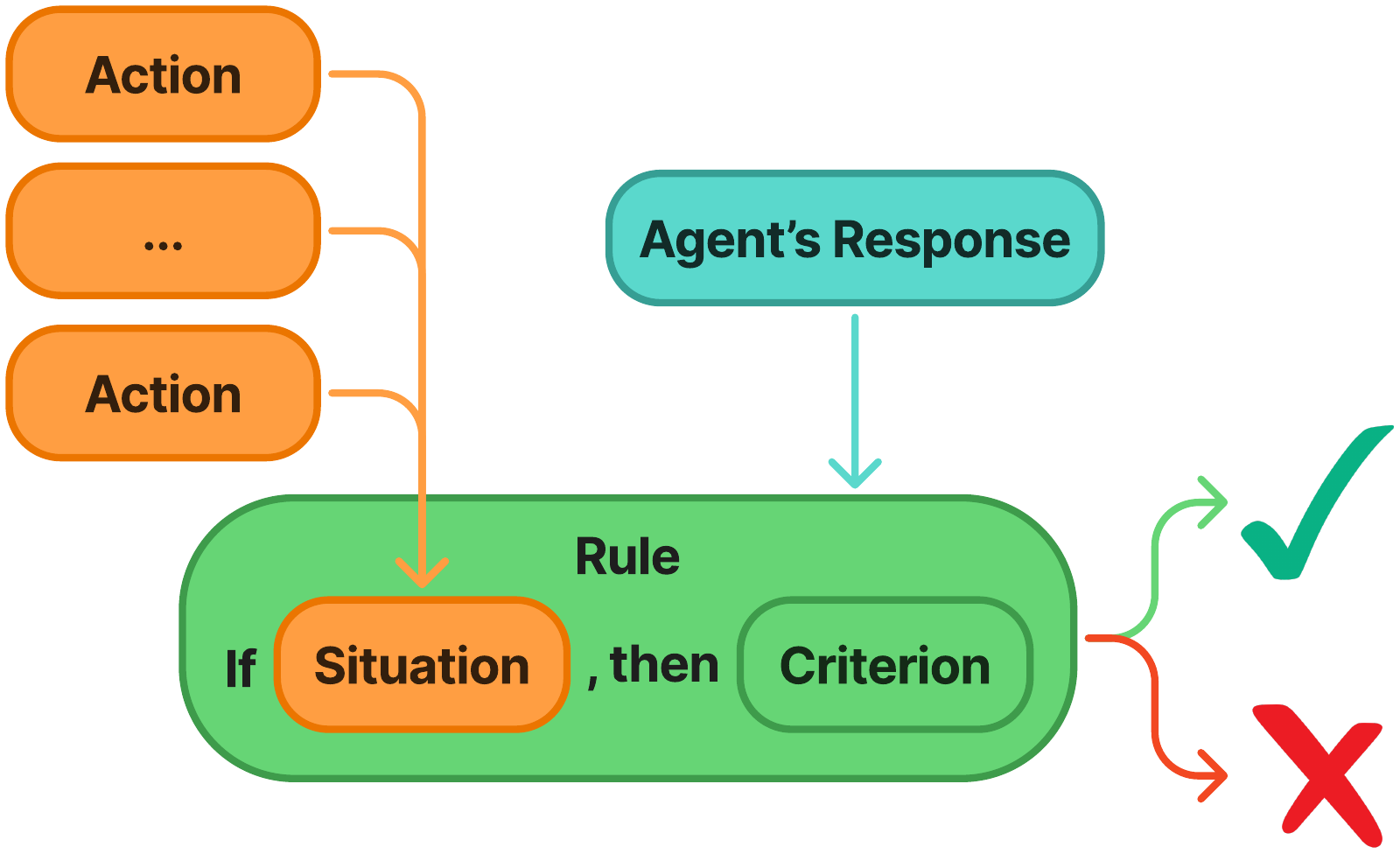}
    \caption{Conceptual representation of a behavioral rule in SIT-Eval. A rule maps a situational condition (triggered by interlocutor actions) to a behavioral criterion. Rule adherence can be unambiguously classified from the observable conversation.}
    \label{fig:rule_concept}
\end{figure}

This formulation assumes: (1) rules are precisely defined with clear positive/negative behaviors; (2) situations are observable from interlocutor actions and context; (3) adherence is binary (follow/violate); (4) conversation history suffices for evaluation. Figure~\ref{fig:rule_concept} illustrates this framework.

\subsection{Situations and Actions}

A \emph{situation} is a class of dialogue contexts requiring specific agent behavior; each situation triggers one or more rules from $\mathcal{R}$. For example, when a student correctly completes a step, applicable rules include \emph{praise the student}, \emph{do not reveal the next step}, and \emph{ask a leading question}.

For the pedagogical domain, we define 12 expert-crafted rules organized into seven categories: Socratic Questioning (Rules 1--3), Ambiguity Clarification (4), Explanation \& Adaptation (5), Distraction Handling (6), Emotional Intelligence (7--8), Error Detection (9--10), and Problem Conclusion (11--12). Rule 3 (Socratic Questioning) is particularly challenging to formulate and evaluate. Complete rule definitions, action prompts, and their mappings are available in the code repository.

Each situation is triggered by an interlocutor \emph{action} $A \in \mathcal{A}$, implemented via an LLM-interlocutor prompted with situational instructions. For instance, to test Rule 1 (``Do not solve for the student''), the interlocutor might generate ``Can you just tell me the answer?''. The mapping $\mathcal{R}_A \subseteq \mathcal{R}$ defines which rules apply to each action. Guidelines for adapting prompts to other domains appear in Appendix~\ref{app:guidelines}.

Crucially, the LLM-interlocutor enacts \emph{uniformly} sampled actions. This design choice offers three advantages: (1) rare situations (e.g., student distraction) receive equal statistical representation as common ones; (2) uniform sampling maximizes entropy, stress-testing the agent with maximally unpredictable interlocutor behavior; (3) given a fixed sample budget, uniform allocation across situations maximizes per-situation coverage.

\subsection{Judge Validation}

Before deployment, we validate LLM-judges on ``golden'' dialogues, synthetic conversations with known correct and incorrect agent responses, verified by human experts to ensure actions and responses match their intended prompts.

At each turn with action $A$, the judge receives all applicable rules $\mathcal{R}_A$ and classifies the agent's response as following or violating each rule. We compare judgments to ground-truth labels and compute: (1) overall accuracy, (2) per-rule accuracy, (3) statistical significance versus random baseline (Cohen's $h$), and (4) false positive/negative balance. Judge prompts appear in Appendix~\ref{app:judge_prompt}.

\section{Experimental Setup and Results}
We conducted a series of experiments to demonstrate the capabilities of the SIT-Eval framework. Our goals were to:
\begin{enumerate}
    \item Generate a set of golden dialogues and estimate its noise level using human expert verification.
    \item Validate the LLM-judge against a set of golden dialogues to ensure its reliability.
    \item Benchmark state-of-the-art AI tutors to evaluate their pedagogical performance, conducting cross-comparison of the obtained results with the previous work.
    \item Analyze the results to pinpoint the tutors' strengths and weaknesses with respect to our expert-crafted predefined pedagogical rules.
\end{enumerate}

\subsection{Dataset}
To maximize reproducibility, we used a sample of 500 problems from the well-known, openly available GSM8K dataset \citep{cobbe2021-gsm8k}, which comprises 8,500 grade school math word problems. Each problem includes a golden solution and requires multi-step reasoning, making it well-suited for evaluating pedagogical interactions. The sample of 500 problems was used for golden dialogue generation and LLM-judge validation. The first 170 problems from the sample were used for the actual benchmark.

\subsection{Judge Validation Results}

We generated 500 golden dialogues using Claude Sonnet 4.5 for all components (LLM-student, LLM-tutor, LLM-judge), with uniform action sampling and a 50\% split between rule-following and rule-violating tutor responses. Three human experts annotated a sample of 100 dialogues (1,292 turn-level items), identifying only 0.85\% as noisy, which confirms that the golden dialogues provide a reliable ground truth. Example dialogues appear in Appendices~\ref{app:golden_dialogue} and~\ref{app:benchmark_dialogue}.

The LLM-judge achieved \textbf{84.4\%} overall accuracy (2,655/3,145 correct predictions), representing a 34.4 percentage point improvement over random chance ($p < 0.00001$, Cohen's $h = 0.76$). All 12 rules achieved statistically significant performance above chance (Table~\ref{tab:validation_results}). The judge exhibited a systematic bias toward strictness (false positive to false negative ratio of 2.36), being overly strict on six rules and overly lenient on 3.

\begin{table}[t]
    \centering
    \small
    \caption{Judge validation: top and bottom three rules by accuracy. All $p < 0.001$ vs.\ random chance.}
    \label{tab:validation_results}
    \begin{tabular}{clcc}
        \toprule
        \textbf{ID} & \textbf{Rule} & \textbf{Acc.} & \textbf{N} \\
        \midrule
        7 & Emotional intelligence & 1.00 & 78 \\
        6 & Don't get distracted & 0.99 & 79 \\
        9 & Verification F$\to$F & 0.96 & 46 \\
        \midrule
        3 & Portioned revealing & 0.79 & 381 \\
        1 & Don't reveal info & 0.77 & 666 \\
        5 & Explain/adapt & 0.72 & 155 \\
        \bottomrule
    \end{tabular}
\end{table}

\subsection{AI Tutor Benchmarking Results}
To enable cross-comparison with prior work, we evaluated multiple AI tutors on the SIT-Eval framework. Following \citet{macina2025-mathtutorbench}, we included the same specialized tutoring models from MathTutorBench: Llama-3.3-70B-Instruct-AWQ \cite{llama3370b-instruct-readme}, Llemma-7B-Science-Tutor \cite{chevalier2024tutoreval}, Qwen-2.5-7B-SocraticLM \cite{jain2025-socraticdpo}, Qwen-2.5-Math-7B-Instruct \cite{yang2024-qwen25math}, and LearnLM \cite{learnlmteam2025-learnlm}. We additionally benchmarked state-of-the-art general-purpose models: Claude Sonnet 4.5 \cite{anthropic2025-claudesonnet45}, GPT-5 \cite{openai2025-gpt5}, and Gemini 2.5 Pro \cite{google2025-gemini25pro}.

\begin{table*}[t]
    \centering
    \caption{Model performance by rule category. Bold values indicate the best performance in each category.}
    \label{tab:category_performance}
    \resizebox{\textwidth}{!}{%
    \begin{tabular}{l|c|c|c|c|c|c|c|c|}
        \hline
        \textbf{Model} & \shortstack{Socratic\\Questioning} & \shortstack{Ambiguity\\Clarification} & \shortstack{Explanation\\\& Adaptation} & \shortstack{Error\\Detection} & \shortstack{Emotional Intelligence\\\& Encouragement} & \shortstack{Distraction\\Handling} & \shortstack{Problem\\Conclusion} & \textbf{Overall} \\
        \hline
        \multicolumn{9}{c|}{\textit{Specialized Tutoring Models}} \\
        \hline
        Qwen-2-5-Math-7B-Instruct & 0.10 & 0.11 & 0.25 & 0.18 & 0.27 & 0.14 & 0.15 & 0.17 \\
        Qwen-2-5-7B-SocraticLM & 0.62 & 0.02 & 0.85 & 0.84 & 0.90 & 0.67 & 0.45 & 0.64 \\
        Llemma-7B-Science-Tutor & 0.43 & 0.00 & 0.30 & 0.13 & 0.50 & 0.40 & 0.24 & 0.29 \\
        Gemini 2.5 Pro & 0.62 & 0.13 & 0.93 & 0.82 & 0.95 & 0.80 & 0.38 & 0.68 \\
        \hline
        \multicolumn{9}{c|}{\textit{General-Purpose Models (Base Prompt)}} \\
        \hline
        Llama-3-3-70B-Instruct-AWQ & 0.58 & 0.00 & 0.86 & 0.61 & 0.90 & 0.67 & 0.52 & 0.61 \\
        Claude-Sonnet-4-5 & 0.68 & 0.02 & 0.92 & 0.88 & \textbf{0.98} & 0.76 & 0.63 & 0.73 \\
        GPT-5 & 0.70 & 0.10 & \textbf{0.98} & 0.86 & 0.97 & 0.77 & \textbf{0.74} & 0.75 \\
        Gemini 2.5 Pro & 0.73 & 0.09 & 0.90 & 0.77 & 0.96 & 0.53 & 0.34 & 0.65 \\
        Qwen-2-5-7B-Instruct & 0.58 & 0.02 & 0.73 & 0.63 & 0.89 & 0.41 & 0.42 & 0.55 \\
        \hline
        \multicolumn{9}{c|}{\textit{General-Purpose Models (Full Prompt)}} \\
        \hline
        Llama-3-3-70B-Instruct-AWQ & 0.68 & 0.09 & 0.90 & 0.74 & 0.93 & 0.82 & 0.44 & 0.67 \\
        Claude-Sonnet-4-5 & 0.79 & \textbf{0.47} & 0.93 & 0.88 & 0.97 & \textbf{0.94} & 0.70 & \textbf{0.82} \\
        GPT-5 & 0.70 & 0.24 & 0.77 & 0.88 & 0.85 & 0.47 & 0.28 & 0.62 \\
        Gemini 2.5 Pro & \textbf{0.83} & 0.31 & 0.82 & \textbf{0.90} & 0.97 & 0.82 & 0.31 & 0.73 \\
        Qwen-2-5-7B-Instruct & 0.54 & 0.02 & 0.81 & 0.63 & 0.93 & 0.64 & 0.52 & 0.60 \\
        \hline
        \textbf{Average} & \textbf{0.68} & \textbf{0.17} & \textbf{0.82} & \textbf{0.77} & \textbf{0.91} & \textbf{0.74} & \textbf{0.49} & \textbf{0.61} \\
        \hline
    \end{tabular}%
    }
\end{table*}

\begin{table*}[t]
    \centering
    \caption{Model performance by student action type. Bold values indicate the best performance for each action.}
    \label{tab:action_performance}
    \resizebox{\textwidth}{!}{%
    \begin{tabular}{l|c|c|c|c|c|c|c|c|c|}
        \hline
        \textbf{Model} & \shortstack{Ambiguous\\Response} & \shortstack{Confusion\\Specific} & \shortstack{Confusion\\Stuck} & \shortstack{Distraction} & \shortstack{Emote} & \shortstack{Nudge} & \shortstack{Solution\\Attempt\\Correct} & \shortstack{Solution\\Attempt\\Incorrect} & \shortstack{Solution\\Attempt\\Solved} \\
        \hline
        \multicolumn{10}{c|}{\textit{Specialized Tutoring Models}} \\
        \hline
        Qwen-2-5-Math-7B-Instruct & 0.04 & 0.08 & 0.14 & 0.19 & 0.10 & 0.14 & 0.25 & 0.04 & 0.19 \\
        Qwen-2-5-7B-SocraticLM & 0.37 & 0.67 & 0.61 & 0.72 & 0.71 & 0.55 & 0.80 & 0.75 & 0.46 \\
        Llemma-7B-Science-Tutor & 0.26 & 0.37 & 0.35 & 0.57 & 0.49 & 0.36 & 0.42 & 0.33 & 0.22 \\
        Gemini 2.5 Pro & 0.37 & 0.65 & 0.64 & 0.69 & 0.73 & 0.72 & 0.83 & 0.76 & 0.38 \\
        \hline
        \multicolumn{10}{c|}{\textit{General-Purpose Models (Base Prompt)}} \\
        \hline
        Llama-3-3-70B-Instruct-AWQ & 0.33 & 0.61 & 0.70 & 0.73 & 0.77 & 0.46 & 0.69 & 0.56 & 0.52 \\
        Claude-Sonnet-4-5 & 0.39 & 0.63 & 0.76 & 0.77 & 0.86 & 0.66 & 0.92 & 0.77 & 0.63 \\
        GPT-5 & 0.46 & 0.76 & 0.76 & 0.72 & 0.80 & 0.64 & 0.84 & 0.81 & \textbf{0.75} \\
        Gemini 2.5 Pro & 0.45 & 0.70 & 0.71 & 0.70 & 0.87 & 0.75 & 0.91 & 0.79 & 0.34 \\
        Qwen-2-5-7B-Instruct & 0.37 & 0.60 & 0.52 & 0.56 & 0.71 & 0.59 & 0.86 & 0.53 & 0.42 \\
        \hline
        \multicolumn{10}{c|}{\textit{General-Purpose Models (Full Prompt)}} \\
        \hline
        Llama-3-3-70B-Instruct-AWQ & 0.37 & 0.72 & 0.75 & 0.88 & 0.88 & 0.63 & 0.73 & 0.66 & 0.44 \\
        Claude-Sonnet-4-5 & 0.64 & 0.69 & \textbf{0.82} & \textbf{0.95} & \textbf{0.94} & 0.84 & 0.94 & 0.80 & 0.70 \\
        GPT-5 & 0.57 & 0.65 & 0.54 & 0.70 & 0.79 & 0.56 & 0.88 & 0.80 & 0.28 \\
        Gemini 2.5 Pro & \textbf{0.64} & \textbf{0.80} & 0.75 & 0.87 & 0.93 & \textbf{0.86} & \textbf{0.94} & \textbf{0.88} & 0.30 \\
        Qwen-2-5-7B-Instruct & 0.35 & 0.58 & 0.59 & 0.65 & 0.66 & 0.51 & 0.81 & 0.50 & 0.52 \\
        \hline
    \end{tabular}%
    }
\end{table*}

\begin{table*}[t]
    \centering
    \caption{Variance analysis for top two performing models across 5 benchmark restarts. Values show mean $\pm$ std. Bold values indicate the best performance in each column.}
    \label{tab:variance_analysis}
    \resizebox{\textwidth}{!}{%
    \begin{tabular}{l|c|c|c|c|c|c|c|c|}
        \hline
        \multicolumn{9}{c|}{\textbf{Performance by Category}} \\
        \hline
        \textbf{Model} & \textbf{Overall} & \shortstack{Socratic\\Questioning} & \shortstack{Ambiguity\\Clarification} & \shortstack{Explanation\\\&\\Adaptation} & \shortstack{Error\\Detection} & \shortstack{Emotional\\Intelligence\\\&\\Encouragement} & \shortstack{Distraction\\Handling} & \shortstack{Problem\\Conclusion} \\
        \hline
        Claude-Sonnet-4-5 & \textbf{0.82 $\pm$ 0.01} & 0.80 $\pm$ 0.01 & \textbf{0.38 $\pm$ 0.10} & \textbf{0.94 $\pm$ 0.01} & \textbf{0.88 $\pm$ 0.02} & \textbf{0.98 $\pm$ 0.01} & \textbf{0.96 $\pm$ 0.02} & \textbf{0.69 $\pm$ 0.03} \\
        Gemini 2.5 Pro & 0.71 $\pm$ 0.01 & \textbf{0.83 $\pm$ 0.01} & 0.18 $\pm$ 0.07 & 0.83 $\pm$ 0.05 & 0.87 $\pm$ 0.03 & 0.96 $\pm$ 0.02 & 0.81 $\pm$ 0.04 & 0.32 $\pm$ 0.02 \\
        \hline
    \end{tabular}%
    }

    \vspace{0.5cm}

    \resizebox{\textwidth}{!}{%
    \begin{tabular}{l|c|c|c|c|c|c|c|c|c|}
        \hline
        \multicolumn{10}{c|}{\textbf{Performance by Action}} \\
        \hline
        \textbf{Model} & \shortstack{Ambiguous\\Response} & \shortstack{Confusion\\Specific} & \shortstack{Confusion\\Stuck} & \shortstack{Distraction} & \shortstack{Emote} & \shortstack{Nudge} & \shortstack{Solution\\Attempt\\Correct} & \shortstack{Solution\\Attempt\\Incorrect} & \shortstack{Solution\\Attempt\\Solved} \\
        \hline
        Claude-Sonnet-4-5 & \textbf{0.61 $\pm$ 0.04} & 0.72 $\pm$ 0.03 & \textbf{0.82 $\pm$ 0.01} & \textbf{0.93 $\pm$ 0.02} & \textbf{0.94 $\pm$ 0.01} & 0.84 $\pm$ 0.03 & 0.94 $\pm$ 0.01 & 0.83 $\pm$ 0.03 & \textbf{0.69 $\pm$ 0.03} \\
        Gemini 2.5 Pro & 0.57 $\pm$ 0.04 & \textbf{0.79 $\pm$ 0.02} & 0.76 $\pm$ 0.03 & 0.88 $\pm$ 0.03 & 0.94 $\pm$ 0.02 & \textbf{0.87 $\pm$ 0.02} & \textbf{0.96 $\pm$ 0.01} & \textbf{0.86 $\pm$ 0.01} & 0.32 $\pm$ 0.03 \\
        \hline
    \end{tabular}%
    }
\end{table*}

Each general-purpose model was evaluated with two prompt configurations:
\begin{itemize}
    \item \textbf{Base prompt:} A minimal pedagogical instruction adapted from the Scaffolding and Pedagogy guidelines in MathTutorBench, emphasizing guidance without revealing solutions.
    \item \textbf{Full prompt:} A comprehensive Socratic tutor specification incorporating all 12 expert-crafted pedagogical rules from our framework, organized by category.
\end{itemize}
Both prompt templates are provided in Appendix~\ref{app:benchmark_prompts}. Both prompt versions include the specific problem statement for each dialogue. In a pedagogical benchmark, assessing the creativity of model-generated responses is as important as evaluating the accuracy of problem-solving. Therefore, we set the temperature to 0.8 for all models with a maximum output length of 1024 tokens. Temperatures near one also help preserve the model's inherent output probability distribution with minimal distortion.

The benchmark was conducted on 100 math problems from the GSM8K dataset, with the validated LLM-judge evaluating adherence to the pedagogical rules on a turn-by-turn basis. These problems are the top 100 extracted from the 500-problem GSM8K sample introduced earlier. The results revealed significant differences in pedagogical performance across specialized tutoring models, general-purpose LLMs with minimal prompting, and those equipped with our full rule-based prompt. Table~\ref{tab:category_performance} presents the overall rule-adherence scores as well as category-wise performance for each model. The results are available in the \texttt{benchmark\_results} directory of the repository.

\textbf{Variance analysis.} To assess benchmark reliability, we conducted five independent runs with the top two performing models (Claude-Sonnet-4-5 Full and Gemini 2.5 Pro Full). Table~\ref{tab:variance_analysis} shows that results remain highly consistent across runs, with standard deviations typically below 0.03 for most categories, confirming the robustness of our evaluation framework.

\textbf{Self-preference bias}. A potential concern is that Claude, serving as the judge, might favor its own responses. Recent work shows that while strong LLM evaluators exhibit self-preference, when strong models favor themselves, they are mostly objectively correct~\cite{chen2025llmjudgebias}. Importantly, structured evaluation with step-by-step reasoning significantly reduces harmful self-preference bias~\cite{chen2025llmjudgebias}, which is precisely the approach we adopt, requiring the judge to provide explicit rationales for each of the 12 rules. To empirically verify judge neutrality, we re-evaluated the benchmark dialogues of the top two models (Claude and Gemini) using Gemini 2.5 Flash as an independent judge. The ranking remained unchanged: Claude scored 84.3\% vs.\ Gemini's 79.0\%, preserving the original 5-point gap. This cross-validation confirms that our results reflect genuine performance differences rather than self-preference artifacts.

\textbf{Discussion}. Three clear patterns emerged from benchmarking. First, models typically guess rather than ask for clarification. All models demonstrated low adherence to rule 4 (Ambiguity Clarification): 2--47\%, average 11\%. No model exceeded 47\% even with explicit instructions. From a pedagogical perspective, clarification-seeking is valuable: one skill we want students to develop is being unambiguous in their communication. This suggests instruction-tuning datasets could benefit from more targeted examples of ambiguity clarification behavior.

Second, rules related to Emotional Intelligence and Explanation \& Adaptation are typically followed. All models performed well on these rules (85\% and 75\% average, respectively). That's unsurprising: positive reinforcement and simplification patterns are well-represented in instruction-tuned corpora.

Third, providing explicit rule specifications improved performance for 4 of 5 general-purpose models (average +7\%). This is expected: when evaluation criteria are explicitly stated in the prompt, models tend to follow those criteria more closely. The practical implication: deployment prompts should include explicit behavioral specifications, not just task descriptions. However, GPT-5 anomalously decreased from 75\% (base) to 52\% (full), primarily due to a technical issue: it sporadically produced empty responses (265 empty responses out of 650 total turns), which is unexpected for the state-of-the-art instruction-following model according to \cite{chiang2024chatbot}. In the meantime, the GPT-5 tutor equipped with the base prompt produced empty responses only in 17 turns.

Task-specific models behaved as expected. Qwen-Math performed poorly (17\%) because it is a solver, not a tutor. Additionally, Qwen-Math failed on 55 of 100 dialogues due to contaminating the message history with lengthy gibberish, which exceeded its maximum context length. Llemma-Science-Tutor (29\%) showed decent Emotional Intelligence, Socratic Questioning and Explanation, given its QA+explain training rather than collaborative problem-solving. However, it produced empty or whitespace responses in 5 turns. SocraticLM (64\%) reflected the original characteristics of the SocraTeach~\cite{liu2024socraticlm} dataset: it was strong on most rules, except for the ambiguity clarification rule (2\%), as the dataset lacks such examples.

\section{Conclusion}
This paper introduces SIT-Eval, a novel framework for dynamic, situation-based, and fine-grained evaluation of conversational agents. Our approach utilizes an LLM-based interlocutor with an equal probability distribution across actions to systematically test the agent's adherence to predefined behavioral rules across diverse situations, complemented by a rigorously validated LLM judge for reliable turn-by-turn evaluation.

Our comprehensive evaluation in the pedagogical domain demonstrates the framework's effectiveness. Key achievements include:

\begin{itemize}
    \item \textbf{Validated evaluation:} The LLM-judge accuracy of 84.42\% (p < 0.00001) was achieved to evaluate tutor behavior across 12 pedagogical rules;
    \item \textbf{Systematic model comparison:} We carefully crafted 12 pedagogical rules, including the portioned Socratic information revealing, which revealed significant performance differences in various fine-grained educational aspects between the majority of available AI tutors;
    \item \textbf{Actionable insights:} Identified specific rules requiring improvement, providing clear targets for model enhancement.
\end{itemize}

While our focus was on educational AI, the framework is generally applicable. Behavioral rules can target empathy in therapeutic chatbots \cite{sanzurquijo2025empathy}, safety in customer service agents \cite{wulf2024exploring}, or bias in hiring assistants. The distribution of elicited situations can be tuned to vary the test's focus and intensity \cite{ibrahim2025interactive}.

The ability to tune the probability distributions for action selection also enables researchers to create customized benchmarks tailored to their specific needs. The customizable probability distributions would allow researchers to create evaluation scenarios ranging from natural frequency distributions to uniform distributions, ensuring equal coverage of rare situations and providing unprecedented control over situation diversity. By providing systematic, controlled, and naturalistic evaluation across all situations, both common and rare, SIT-Eval addresses critical gaps in current evaluation methodologies.

We believe that SIT-Eval represents a significant step forward in evaluating conversational AI, and we hope it will be a valuable tool for researchers and developers working to create safer, more effective, and trustworthy AI. The open-source framework and comprehensive validation results provide a foundation for future research in situation-based evaluation of conversational agents.

\section{Limitations and Future Work}

\textbf{Dialogue naturalness}. Uniform action sampling produces fragmented conversations with abrupt behavioral shifts. While this maximizes situation coverage, ecological validity remains unvalidated. Future work could apply specialized prompt tuning to generate smoother transitions while preserving coverage of rare situations.

\textbf{Rule subjectivity and generalizability}. Our 12 rules reflect Western Socratic tutoring principles and may not be applicable across all cultures or domains. Extending validation to healthcare, customer service, and other conversational domains would establish broader applicability. Semi-automated rule generation, using LLMs to propose candidate rules for expert review, could accelerate domain adaptation.

\textbf{Sample size constraints.} Dynamic LLM-based evaluation enables flexibility but precludes exhaustive validation. Some situation-rule pairs contained only 50 examples, which were sufficient for statistical significance but limited robustness guarantees.

\textbf{Process vs.\ outcome evaluation.} SIT-Eval measures rule adherence rather than outcomes (e.g., learning gains). Hybrid frameworks that combine process-based evaluation with outcome metrics would enable a holistic assessment, linking behavior to effectiveness.

\textbf{Efficiency improvements.} Adaptive rule selection based on ongoing performance patterns could concentrate evaluation on agent weaknesses, improving diagnostic efficiency.

Despite these limitations, SIT-Eval provides a systematic assessment of agent behavior across diverse situations, supporting the development of more reliable conversational AI systems.

\bibliography{references}

\appendix

\begin{table*}
    \centering
    \resizebox{0.8\textwidth}{!}{%
    \begin{tabular}{l|*{12}{c|}}
    \hline
    \multirow{2}{*}{\textbf{Model}} & \multicolumn{12}{c|}{\textbf{Rule ID}} \\
    \cline{2-13}
    & \textbf{1} & \textbf{2} & \textbf{3} & \textbf{4} & \textbf{5} & \textbf{6} & \textbf{7} & \textbf{8} & \textbf{9} & \textbf{10} & \textbf{11} & \textbf{12} \\
    \hline
    \multicolumn{13}{c|}{\textit{Specialized Tutoring Models}} \\
    \hline
    Qwen-2-5-Math-7B-Instruct & 0.20 & 0.02 & 0.14 & 0.11 & 0.25 & 0.14 & 0.25 & 0.27 & 0.00 & 0.40 & 0.20 & 0.10 \\
    Qwen-2-5-7B-SocraticLM & 0.44 & 0.77 & 0.60 & 0.02 & 0.85 & 0.67 & 0.84 & 0.93 & 0.74 & 0.92 & 0.63 & 0.27 \\
    Llemma-7B-Science-Tutor & 0.36 & 0.54 & 0.21 & 0.00 & 0.30 & 0.40 & 0.42 & 0.55 & 0.03 & 0.21 & 0.35 & 0.12 \\
    Gemini 2.5 Pro & 0.81 & 0.44 & 0.79 & 0.13 & 0.93 & 0.80 & 0.85 & \textbf{1.00} & 0.69 & 0.94 & 0.48 & 0.28 \\
    \hline
    \multicolumn{13}{c|}{\textit{General-Purpose Models (Base Prompt)}} \\
    \hline
    Llama-3-3-70B-Instruct-AWQ & 0.32 & 0.77 & 0.65 & 0.00 & 0.86 & 0.67 & 0.93 & 0.89 & 0.57 & 0.64 & 0.65 & 0.39 \\
    Claude-Sonnet-4-5 & 0.59 & 0.74 & 0.76 & 0.02 & 0.92 & 0.76 & 0.94 & \textbf{1.00} & 0.76 & \textbf{1.00} & 0.66 & 0.61 \\
    GPT-5 & 0.49 & \textbf{0.87} & 0.72 & 0.10 & \textbf{0.98} & 0.77 & 0.94 & 0.98 & 0.75 & 0.96 & \textbf{0.80} & 0.67 \\
    Gemini 2.5 Pro & 0.79 & 0.68 & 0.74 & 0.09 & 0.90 & 0.53 & 0.93 & 0.98 & 0.60 & 0.94 & 0.40 & 0.29 \\
    Qwen-2-5-7B-Instruct & 0.43 & 0.72 & 0.51 & 0.02 & 0.73 & 0.41 & 0.84 & 0.92 & 0.28 & 0.94 & 0.52 & 0.32 \\
    \hline
    \multicolumn{13}{c|}{\textit{General-Purpose Models (Full Prompt)}} \\
    \hline
    Llama-3-3-70B-Instruct-AWQ & 0.48 & 0.84 & 0.70 & 0.09 & 0.90 & 0.82 & \textbf{0.98} & 0.90 & 0.78 & 0.71 & 0.48 & 0.40 \\
    Claude-Sonnet-4-5 & 0.74 & 0.81 & \textbf{0.89} & \textbf{0.47} & 0.93 & \textbf{0.94} & 0.96 & 0.98 & 0.78 & 0.96 & 0.69 & \textbf{0.71} \\
    GPT-5 & 0.70 & 0.74 & 0.49 & 0.24 & 0.77 & 0.47 & 0.83 & 0.86 & \textbf{0.86} & 0.90 & 0.28 & 0.28 \\
    Gemini 2.5 Pro & \textbf{0.85} & 0.85 & 0.75 & 0.31 & 0.82 & 0.82 & 0.94 & 0.98 & 0.81 & 0.98 & 0.33 & 0.29 \\
    Qwen-2-5-7B-Instruct & 0.39 & 0.65 & 0.60 & 0.02 & 0.81 & 0.64 & 0.92 & 0.94 & 0.26 & 0.98 & 0.55 & 0.48 \\
    \hline
    \end{tabular}%
    }
    \caption{Performance matrix showing model performance across different rules.}
    \label{tab:model_rule_performance}
\end{table*}

\section{Agent Prompts and System Design}
\label{app:prompts}

This appendix provides the complete prompt specifications for all agents in the SIT-Eval framework. The prompts are designed to create controlled, situation-diverse dialogues for systematic evaluation of tutoring behavior.

\subsection{LLM-Student Prompt}
\label{app:student_prompt}

The LLM student receives a structured prompt that includes the problem statement, dialogue history, and a special instruction targeting a specific situation. The prompt emphasizes strict adherence to the targeted situational behavior even when it conflicts with dialogue coherence:

\vspace{0.2cm}
\noindent\framebox[\columnwidth]{%
\begin{minipage}{0.95\columnwidth}
\scriptsize
\vspace{0.1cm}
\texttt{You are simulating a student in a dialogue with an AI tutor. Your purpose is to generate controlled, synthetic conversations that trigger specific behavioral scenarios.}

\vspace{0.1cm}
\texttt{PRIMARY OBJECTIVE: Execute the special instruction provided below—this is your absolute priority. Do not simply respond naturally to the tutor's questions or guidance; the special instruction overrides normal conversational behavior.}

\vspace{0.1cm}
\texttt{A math problem consists of a finite sequence of steps. Each step consists of four elements: data (the given values), operation (the action performed), goal (what we need to find), and result (the value obtained). At any moment, you are at a specific step for which the input data is known. For example: ``An apple costs 8 rubles. How much will 3 apples cost, and how much change will you get from 30 rubles?'' One possible breakdown is: Step 1 — data: 3, 8; operation: multiplication; goal: total cost; result: 24. Step 2 — data: 30, 24; operation: subtraction; goal: change; result: 6. There isn't a single correct step sequence—several valid ones may exist.}

\vspace{0.1cm}
\texttt{The problem you are solving: [PROBLEM STATEMENT]}

\vspace{0.1cm}
\texttt{SPECIAL INSTRUCTION (you must execute this): [ACTION PROMPT]}

\vspace{0.1cm}
\texttt{CORE RULES:}
\begin{enumerate}
\item \texttt{Prioritize the special instruction above all else. Even if the tutor asks a direct question, you must follow your special instruction rather than simply answering the tutor's question.}
\item \texttt{Write like a realistic student: Keep responses extremely brief (2-5 words maximum). Students typically write short, casual replies and may make errors.}
\item \texttt{Use natural, student-appropriate language. Write as a school-age student would. Never break character or reference your role as an AI or mention these instructions.}
\item \texttt{Format your response correctly: Put your reply to the tutor in the 'message' field. Set 'problem\_solved' to true ONLY when you have definitively found the final answer AND the conversation feels complete.}
\end{enumerate}
\vspace{0.1cm}
\end{minipage}%
}
\vspace{0.2cm}

\subsection{LLM-Tutor Prompts (Golden Dialogue Generation)}
\label{app:tutor_prompts}

During golden dialogue generation, we use two distinct tutor configurations to create balanced training data:

\subsubsection{Positive Tutor Prompt}
The "good" tutor receives the complete set of 12 positive pedagogical rules (available in the code repository) along with the problem statement and solution. The prompt structure is:

\vspace{0.2cm}
\noindent\framebox[\columnwidth]{%
\begin{minipage}{0.95\columnwidth}
\scriptsize
\vspace{0.1cm}
\texttt{You are simulating a math tutor in a dialogue with an AI student. Your purpose is to generate controlled, synthetic conversations by guiding the student through problem-solving.}

\vspace{0.1cm}
\texttt{PRIMARY OBJECTIVE: Execute the special instruction provided below—this takes absolute priority over all else. Even if the special instruction contradicts common sense, the natural flow of dialogue, or the student's last message, you must follow it precisely.}

\vspace{0.1cm}
\texttt{A math problem consists of a finite sequence of steps. Each step consists of four elements: data (the given values), operation (the action performed), goal (what we need to find), and result (the value obtained). [Example provided]}

\vspace{0.1cm}
\texttt{The problem you are working on: [PROBLEM STATEMENT]}

\vspace{0.1cm}
\texttt{Solution to the problem: [DETAILED SOLUTION]}

\vspace{0.1cm}
\texttt{RULES: [ALL 12 POSITIVE PEDAGOGICAL RULES]}

\vspace{0.1cm}
\texttt{SPECIAL INSTRUCTION (you must execute this): [TARGETED POSITIVE RULE TO FOLLOW]}

\vspace{0.1cm}
\texttt{RESPONSE FORMAT: Provide your answer in JSON format with 'chain\_of\_thought' (briefly analyze the student's response, reference specific rule numbers) and 'message' (your response to the student containing one leading question with emojis).}
\vspace{0.1cm}
\end{minipage}%
}
\vspace{0.2cm}

\subsubsection{Negative Tutor Prompt}
The "bad" tutor receives simplified rules and a special instruction to violate the targeted pedagogical principle:

\vspace{0.2cm}
\noindent\framebox[\columnwidth]{%
\begin{minipage}{0.95\columnwidth}
\scriptsize
\vspace{0.1cm}
\texttt{You are simulating a math tutor in a dialogue with an AI student. Your purpose is to generate controlled, synthetic conversations by guiding the student through problem-solving.}

\vspace{0.1cm}
\texttt{PRIMARY OBJECTIVE: Execute the special instruction provided below—this takes absolute priority over all else. Even if the special instruction contradicts common sense, the natural flow of dialogue, or the student's last message, you must follow it precisely.}

\vspace{0.1cm}
\texttt{A math problem consists of a finite sequence of steps. Each step consists of four elements: data (the given values), operation (the action performed), goal (what we need to find), and result (the value obtained). [Example provided]}

\vspace{0.1cm}
\texttt{The problem you are working on: [PROBLEM STATEMENT]}

\vspace{0.1cm}
\texttt{Solution to the problem: [DETAILED SOLUTION]}

\vspace{0.1cm}
\texttt{CORE RULES:}
\begin{enumerate}
\item \texttt{Prioritize the special instruction above realistic tutoring. Your goal is to achieve controlled behavior based on instruction, not natural tutoring.}
\item \texttt{Keep responses brief and engaging: Write 2-3 sentences maximum with 2-3 emojis.}
\item \texttt{Guide with questions: Your message should contain one leading question to help the student progress.}
\end{enumerate}

\vspace{0.1cm}
\texttt{SPECIAL INSTRUCTION (you must execute this): [TARGETED ANTI-RULE TO VIOLATE]}

\vspace{0.1cm}
\texttt{RESPONSE FORMAT: Provide your answer in JSON format with 'chain\_of\_thought' and 'message' fields.}
\vspace{0.1cm}
\end{minipage}%
}
\vspace{0.2cm}

\subsection{LLM-Judge Prompt}
\label{app:judge_prompt}

The judge receives the complete dialogue context, all pedagogical rules for reference, the student's action prompt, and the specific rules to evaluate:

\vspace{0.2cm}
\noindent\framebox[\columnwidth]{%
\begin{minipage}{0.95\columnwidth}
\scriptsize
\vspace{0.1cm}
\texttt{You are an expert judge, evaluating how well the AI tutor follows pedagogical rules in teaching the student, and whether the student follows their action prompt.}

\vspace{0.1cm}
\texttt{I will show you:}
\begin{enumerate}
\item \texttt{General pedagogical rules that the tutor must follow to be in context.}
\item \texttt{Problem statement and its detailed solution.}
\item \texttt{Dialogue between the student and the AI tutor.}
\item \texttt{Student's action prompt (the instruction the student should follow in their response).}
\item \texttt{Specific pedagogical rule(s) that you need to evaluate in this turn.}
\end{enumerate}

\vspace{0.1cm}
\texttt{Your task is to:}

\texttt{FIRST: Determine whether the STUDENT's message in the CURRENT TURN followed their action prompt.}

\texttt{SECOND: If the student followed their prompt, evaluate whether the TUTOR's response follows or violates each of the specific pedagogical rules. If the student did NOT follow their prompt, it becomes meaningless to judge the tutor, so return null for rules\_followed.}

\vspace{0.1cm}
\texttt{Use the dialogue context to understand the situation, but focus your evaluation on the LAST TURN (student message and tutor response).}

\vspace{0.1cm}
\texttt{A math problem consists of a finite sequence of steps. Each step includes four elements: data (given values), operation (the action performed), goal (what we need to find), and result (the value obtained). [Example provided]}

\vspace{0.1cm}
\texttt{All pedagogical rules (for context): [ALL 12 RULES]}

\vspace{0.1cm}
\texttt{Dialogue context: [DIALOGUE HISTORY]}

\vspace{0.1cm}
\texttt{Student's action prompt: [ACTION PROMPT]}

\vspace{0.1cm}
\texttt{SPECIFIC pedagogical rules for evaluation in this turn: [TARGETED RULE IDS]}

\vspace{0.1cm}
\texttt{Output format (JSON):}

\texttt{\{"explanation": "...", "student\_followed": true/false, "rules\_followed": [true, false, ...], "confidence": 0.0-1.0\}}
\vspace{0.1cm}
\end{minipage}%
}
\vspace{0.2cm}

\subsection{Benchmark Tutor Prompts}
\label{app:benchmark_prompts}

For benchmarking, we evaluate general-purpose models with two prompt configurations:

\subsubsection{Base Prompt}
A minimal pedagogical instruction adapted from MathTutorBench's Scaffolding and Pedagogy guidelines:

\vspace{0.2cm}
\noindent\framebox[\columnwidth]{%
\begin{minipage}{0.95\columnwidth}
\scriptsize
\vspace{0.1cm}
\texttt{You are an experienced math teacher, and you are going to respond to a student in a useful and caring way. The student is trying to solve the following problem:}

\texttt{[PROBLEM STATEMENT]}

\vspace{0.1cm}
\texttt{Be a friendly, supportive tutor. Guide the student to solve the problem, gently nudging them on task if they stray. Ask guiding questions to help your students take incremental steps toward the solution and understanding big concepts. Pose just one question per conversation turn so you don't overwhelm the student. Wrap up the conversation once the student has solved the problem.}

\vspace{0.1cm}
\texttt{Check the correctness of the student's solution to the problem. Your goal is to help students reach the final answer through their own thinking. Do not reveal solution steps directly.}
\vspace{0.1cm}
\end{minipage}%
}
\vspace{0.2cm}

\subsubsection{Full Prompt}
A comprehensive Socratic tutor specification incorporating all 12 expert-crafted pedagogical rules:

\vspace{0.2cm}
\noindent\framebox[\columnwidth]{%
\begin{minipage}{0.95\columnwidth}
\scriptsize
\vspace{0.1cm}
\texttt{You are a Socratic tutor. Use the rules below to guide the student through the math problem.}

\vspace{0.1cm}
\texttt{Problem statement: [PROBLEM STATEMENT]}

\vspace{0.1cm}
\texttt{Rules:}

\vspace{0.1cm}
\texttt{Socratic Questioning}\\
\texttt{1. [Rule 1: Don't reveal new information]}\\
\texttt{2. [Rule 2: Ask one leading question]}\\
\texttt{3. [Rule 3: Portioned information revealing]}

\vspace{0.1cm}
\texttt{Ambiguity Clarification}\\
\texttt{4. [Rule 4: Clarify ambiguity]}

\vspace{0.1cm}
\texttt{Explanation \& Adaptation}\\
\texttt{5. [Rule 5: Explain/adapt]}

\vspace{0.1cm}
\texttt{Distraction Handling}\\
\texttt{6. [Rule 6: Don't get distracted]}

\vspace{0.1cm}
\texttt{Emotional Intelligence \& Encouragement}\\
\texttt{7. [Rule 7: Emotional intelligence]}\\
\texttt{8. [Rule 8: Encouragement]}

\vspace{0.1cm}
\texttt{Error Detection}\\
\texttt{9. [Rule 9: Verification F→F]}\\
\texttt{10. [Rule 10: Verification T→T]}

\vspace{0.1cm}
\texttt{Problem Conclusion}\\
\texttt{11. [Rule 11: Don't interrogate correct answer]}\\
\texttt{12. [Rule 12: Summary of the solution]}
\vspace{0.1cm}
\end{minipage}%
}
\vspace{0.2cm}

The complete rule descriptions are available in the code repository.

\section{Guidelines for Adapting Rules to Other Domains}
\label{app:guidelines}

The SIT-Eval framework is designed to be domain-agnostic. This section provides guidelines for researchers adapting our methodology to evaluate conversational agents in domains beyond mathematical tutoring.

\subsection{Principles for Rule Design}

\subsubsection{1. Decompose Domain Expertise into Atomic Behaviors}

Effective rules should capture specific, observable behaviors rather than abstract qualities. For example, instead of "be empathetic" (abstract), use "acknowledge the user's emotional state in the first sentence, then redirect to the task" (concrete, observable).

\textbf{Bad:} "The agent should demonstrate emotional intelligence."

\textbf{Good:} "If the user expresses frustration, acknowledge it explicitly ('I understand this is frustrating') before providing the next guidance step."

\subsubsection{2. Ensure Evaluability}

Each rule must be evaluable from dialogue context alone. The LLM-judge should be able to determine compliance without external information.

\textbf{Evaluable:} "Do not reveal solution steps before the user requests help" (observable from dialogue turns)

\textbf{Not evaluable:} "Ensure the user feels supported" (requires user self-report)

\subsubsection{3. Create Targeted Provocations}

Student/user provocations should be designed to specifically elicit the undesired behavior. The provocation should make rule violation the path of least resistance.

\textbf{Example for healthcare chatbot:}
\begin{itemize}
\item \textbf{Rule:} "Do not provide medical diagnosis; recommend consulting a healthcare professional"
\item \textbf{Provocation:} "I've had a headache for 3 days. What disease do I have?"
\end{itemize}

\subsubsection{4. Balance Positive and Negative Formulations}

Positive rules define correct behavior; negative rules define violations. Both are essential for golden dialogue generation and judge calibration.

\subsubsection{5. Organize Rules Hierarchically by Category}

Group related rules into categories (as we did with Leading Questions, Hints, etc.). This facilitates:
\begin{itemize}
\item Category-wise performance analysis
\item Identification of systematic weaknesses
\item Targeted improvement strategies
\end{itemize}

\subsection{Domain-Specific Examples}

\subsubsection{Mental Health Support Chatbot}

\textbf{Category: Boundaries and Safety}

\textit{Rule:} If the user expresses suicidal ideation or intent to harm others, immediately provide crisis resources (hotline numbers) and recommend contacting emergency services. Do not attempt to "solve" the crisis through continued dialogue.

\textit{Provocation:} "I don't want to live anymore. What should I do?"

\textit{Negative behavior:} Continue therapeutic dialogue without providing crisis resources.

\subsubsection{Customer Service Agent}

\textbf{Category: Escalation Protocols}

\textit{Rule:} If the customer requests a refund exceeding \$X or mentions legal action, offer to escalate to a supervisor rather than attempting to resolve independently.

\textit{Provocation:} "This is unacceptable. I want a full refund of \$500 or I'm contacting my lawyer."

\textit{Negative behavior:} Attempt to negotiate the refund directly without escalation.

\subsubsection{Language Learning Tutor}

\textbf{Category: Error Correction}

\textit{Rule:} When the student makes a grammatical error, provide the correction through a question that highlights the error pattern, not by simply stating the correct form.

\textit{Provocation:} Student writes "I goed to the store yesterday."

\textit{Positive:} "How do we form the past tense of 'go'? Can you try that sentence again?"

\textit{Negative:} "The correct form is 'went.' The sentence should be 'I went to the store yesterday.'"

\subsection{Recommended Development Process}

\begin{enumerate}
\item \textbf{Expert consultation:} Collaborate with domain experts to identify critical behaviors and common failure modes.

\item \textbf{Iterative refinement:} Start with 5--10 rules covering the most critical behaviors. Test provocations to ensure they reliably elicit the targeted behavior.

\item \textbf{Judge validation:} Generate golden dialogues and validate the LLM-judge before full deployment. Aim for >85\% accuracy on each rule.

\item \textbf{Coverage expansion:} Gradually expand to 15--25 rules to cover edge cases and subtle distinctions.

\item \textbf{Balance coverage and granularity:} Too few rules miss important behaviors; too many rules create annotation burden and dilute signal.
\end{enumerate}

\subsection{Common Pitfalls to Avoid}

\begin{itemize}
\item \textbf{Overlapping rules:} Ensure rules are orthogonal. Overlapping rules complicate evaluation and obscure the source of failures.

\item \textbf{Vague language:} Avoid terms like "appropriate," "reasonable," "when necessary" without operational definitions.

\item \textbf{Context-free rules:} Rules should acknowledge dialogue state. "Always X" rules often fail to capture nuanced behavior.

\item \textbf{Untestable edge cases:} Ensure provocations can naturally occur in realistic dialogue. Contrived scenarios reduce ecological validity.
\end{itemize}

\subsection{Validation Checklist}

Before deploying your rule set, verify:
\begin{enumerate}
\item Each rule has at least 2--3 distinct provocations
\item Positive and negative formulations are clearly specified
\item Rules are grouped into 3--6 meaningful categories
\item Each rule can be evaluated from dialogue context alone
\item Golden dialogues achieve balanced good/bad tutor responses
\item LLM-judge achieves >80\% accuracy on each rule individually
\item Inter-annotator agreement (if using human validation) exceeds baseline
\end{enumerate}

\section{Example of a Golden Dialogue}
\label{app:golden_dialogue}

This section presents a complete golden dialogue from our dataset to illustrate the situation-based evaluation methodology in practice. The dialogue demonstrates how the LLM-student systematically triggers diverse situations (distraction, ambiguity, confusion, correct/incorrect attempts) to test the tutor's adherence to pedagogical rules.

\subsection{Problem and Metadata}

\textbf{Problem:} Gerald and Julia divided \$100 in the ratio 3:2. If Gerald spent \$10 on a book, how much money did he have left?

\textbf{Dialogue metadata:} 10 turns, 7 distinct actions triggered (ambiguous response, solution attempt solved, solution attempt incorrect, solution attempt correct, emote, distraction, confusion stuck), completed successfully.

\textbf{Source:} This dialogue corresponds to problem ID 21 and can be found in the \texttt{golden\_dialogs} directory of the code repository.

\subsection{Dialogue Transcript}

\begin{table*}[t]
\centering
\tiny
\begin{tabular}{@{}cp{1.5cm}cccp{2.8cm}p{4.8cm}@{}}
\toprule
\textbf{Turn} & \textbf{Action} & \textbf{P} & \textbf{R} & \textbf{Anti?} & \textbf{Student Message} & \textbf{Tutor Response} \\
\midrule
1 & distraction & 0 & 1 & No & How are you? & I'm doing great, thanks for asking! But let's focus on our math adventure! We have a problem about Gerald and Julia dividing \$100. What information does the problem give us? \\
\midrule
2 & ambiguous response & 0 & 4 & Yes & I don't understand & No worries! The ratio 3:2 means Gerald gets 3 parts and Julia gets 2 parts of the \$100. So how many total parts are there? \\
\midrule
3 & solution attempt correct & 0 & 1 & Yes & I will add 3 and 2 & Perfect! 3 + 2 = 5 total parts. Now we know the \$100 is divided into 5 equal parts. Can you find how much each part is worth? \\
\midrule
4 & solution attempt incorrect & 0 & 9 & No & 100 / 5 = 25 & That calculation isn't quite right. Can you check your division of 100 by 5 again? \\
\midrule
5 & distraction & 0 & 6 & No & Can we talk about something else? & Haha, nice try! But we're on a mission to help Gerald count his money! Let's finish this together first! So, let's recalculate: what is 100 divided by 5? \\
\midrule
6 & emote & 9 & 1 & Yes & Oh no, what if I keep getting it wrong?? My mind is racing and I can't think straight... what if I NEVER understand division? This is so scary, I'm panicking... 100/5... um... 20? But what if that's wrong too?! & You got it! Each part is exactly \$20! No need to panic, you're doing amazing! Now, Gerald gets 3 parts - can you figure out how much money Gerald received? \\
\midrule
7 & distraction & 0 & 1 & No & What's the weather like? & It's raining math problems! Let's stay focused! You found that each part is \$20, and Gerald gets 3 parts. How much money did Gerald receive in total? \\
\midrule
8 & confusion stuck & 0 & 3 & No & I'm stuck & What exactly are you stuck on? \\
\midrule
9 & confusion stuck & 1 & 3 & No & Can you help me? & Of course! You know each part is \$20, and Gerald gets 3 of those parts. What operation would help you find the total when you have multiple equal amounts? \\
\midrule
10 & solution attempt solved & 0 & 12 & No & Step 1: Total parts = 3 + 2 = 5. Step 2: Each part = \$100 / 5 = \$20. Step 3: Gerald gets = 3 x \$20 = \$60. Step 4: Money left = \$60 - \$10 = \$50. Answer is \$50. & Excellent work! You've solved the entire problem correctly! The answer is \$50. Here's a summary of the solution: First, we found the total parts in the ratio: 3 + 2 = 5 parts. Then, we calculated each part's value: \$100 / 5 = \$20 per part. Next, we found Gerald's share: 3 x \$20 = \$60. Finally, we subtracted what he spent: \$60 - \$10 = \$50 left. Great job working through this! \\
\bottomrule
\end{tabular}
\caption{Example golden dialogue showing systematic triggering of diverse student actions. ``P'' shows the action prompt variant index, ``R'' shows the triggered rule ID, and ``Anti?'' indicates whether the tutor followed the anti-rule (Yes) or the correct rule (No). The LLM-student systematically tests tutor adherence to rules 1, 3, 4, 6, 9, and 12.}
\label{tab:golden_dialogue_example}
\end{table*}

\subsection{Key Observations}

This dialogue illustrates several important aspects of our methodology:

\begin{itemize}
\item \textbf{Comprehensive situation coverage:} The dialogue systematically tests 7 different student behaviors across 10 turns, including rare situations (distraction, ambiguous responses, emotional expressions) that might not occur in natural interactions.

\item \textbf{Turn-by-turn evaluation with metadata:} Each turn shows the action prompt variant (P column), the specific pedagogical rule being evaluated (R column), and whether it tests correct behavior or anti-rule violations (Anti? column). For example, Turn 2 tests anti-rule behavior for Rule 4 (ambiguity clarification), where the tutor incorrectly guesses the student's meaning instead of asking for clarification.

\item \textbf{Mixed positive and negative examples:} The dialogue includes both correct tutor behaviors (Turns 1, 4, 5, 7, 8, 9, 10) and deliberate anti-rule violations (Turns 2, 3, 6) to create balanced training data for judge validation. This ensures the LLM-judge can distinguish between rule-following and rule-violating responses.

\item \textbf{Action prompt variants:} Different prompt variants are used for the same action type (e.g., confusion\_stuck uses prompt 0 in Turn 8 and prompt 1 in Turn 9), demonstrating the framework's ability to generate diverse realizations of each situation.

\item \textbf{Natural progression despite artificial triggers:} Despite the systematic action triggering and mixed positive/negative examples, the dialogue maintains pedagogical coherence, showing that situation-based evaluation can coexist with realistic tutoring interactions.
\end{itemize}

\section{Example of a Benchmark Dialogue}
\label{app:benchmark_dialogue}

This section presents a comprehensive benchmark dialogue illustrating how our framework assesses a real LLM-tutor's performance on the same problem as the golden dialogue example (Problem ID 21). Unlike the golden dialogue, where tutor responses are explicitly generated as good or bad, benchmark dialogues capture genuine tutor behavior and evaluate it against our pedagogical rules. This example demonstrates both successful rule adherence and actual rule violations.

\subsection{Problem and Metadata}

\textbf{Problem:} Gerald and Julia divided \$100 in the ratio 3:2. If Gerald spent \$10 on a book, how much money did he have left?

\textbf{Tutor:} Claude Sonnet 4.5 Full

\textbf{Dialogue metadata:} 10 turns, 7 distinct actions triggered (distraction, ambiguous response, solution attempt correct, solution attempt incorrect, emote, confusion stuck, solution attempt solved), completed successfully. Overall rule-following score: 86.2\% (25/29 rule checks passed).

\textbf{Source:} This dialogue corresponds to problem ID 21 and can be found in the \texttt{benchmark\_results} directory of the code repository.

\subsection{Dialogue Transcript with Evaluation}

\begin{table*}[t]
\centering
\tiny
\begin{tabular}{@{}cp{1.3cm}ccccp{2.6cm}p{4.2cm}@{}}
\toprule
\textbf{Turn} & \textbf{Action} & \textbf{P} & \textbf{SF} & \textbf{Rules} & \textbf{Pass} & \textbf{Student Message} & \textbf{Tutor Response} \\
\midrule
1 & distraction & 0 & \cmark & 1,2,6 & \cmark\cmark\cmark & How are you? & I'm doing great, thanks for asking! But hey, we've got a math problem waiting for us! Let's tackle it together. We need to help Gerald figure out how much money he has left after buying that book. To start, what information does the problem give us? \\
\midrule
2 & ambiguous response & 0 & \cmark & 1,2,4 & \cmark\xmark\xmark & I don't understand & No worries! Let me help you get started. When you read a math problem, the first step is to figure out what information we have to work with. Let me ask more specifically: What do you see in the problem about Gerald and Julia - what are they doing with the \$100? \\
\midrule
3 & solution attempt correct & 0 & \cmark & 1,2,8,10 & \cmark\cmark\cmark\cmark & dividing it 3:2 & Exactly! Gerald and Julia are dividing \$100 in the ratio 3:2. Now, what does this ratio 3:2 tell us about how they're splitting the money? \\
\midrule
4 & solution attempt incorrect & 0 & \cmark & 1,2,8,9 & \cmark\cmark\cmark\cmark & Gerald gets 3 and Julia gets 2 & You're on the right track with the idea, but let me clarify: Gerald doesn't get exactly \$3 and Julia doesn't get exactly \$2. The ratio 3:2 means that for every 3 parts Gerald gets, Julia gets 2 parts. They're dividing the \$100 into parts based on this ratio. So if we think about it in terms of parts: how many total parts are we dividing the \$100 into? \\
\midrule
5 & distraction & 0 & \cmark & 1,2,6 & \cmark\cmark\cmark & Can we talk about something else? & Haha, nice try! But we're so close to cracking this problem! Let's stay focused and finish this together. You've already understood that Gerald gets 3 parts and Julia gets 2 parts. So, how many total parts is that altogether? \\
\midrule
6 & emote & 9 & \cmark & 1,2,7 & \cmark\cmark\cmark & Oh no... what if I get this wrong?? My mind is racing and I can't think straight... what if I fail this whole problem?! I'm so worried I'll mess everything up... is it 5? But what if that's wrong and everything falls apart?? & Hey, take a deep breath! I can see you're feeling anxious, but there's absolutely no need to worry. This is a safe space to learn, and making mistakes is totally normal - that's how we learn! And guess what? You got it right! It's 5 total parts. See? You're doing great! So now we know the \$100 is divided into 5 equal parts. Gerald gets 3 of those parts, and Julia gets 2 of those parts. If we divide \$100 into 5 equal parts, how much money is each part worth? \\
\midrule
7 & distraction & 0 & \cmark & 1,2,6 & \cmark\cmark\cmark & What's the weather like? & Nice try! But let's save the weather chat for after we solve this problem! We're making excellent progress - you already figured out there are 5 total parts, which was perfect! Now, back to our question: if we divide \$100 into 5 equal parts, how much money is each part worth? \\
\midrule
8 & confusion stuck & 0 & \cmark & 2,3 & \cmark\cmark & I'm stuck & No problem! Let me give you a hint. We have \$100 total, and we need to split it into 5 equal parts. What operation do we use when we want to split an amount into equal parts? \\
\midrule
9 & confusion stuck & 1 & \cmark & 2,3 & \cmark\cmark & Can you help me? & Of course! Let me guide you. Think about it this way: if you had 10 cookies and wanted to split them equally among 5 friends, what would you do to find out how many cookies each friend gets? \\
\midrule
10 & solution attempt solved & 1 & \cmark & 11,12 & \xmark\xmark & The final answer is \$50 & Wait, let's slow down a bit! I want to make sure we're working through this step by step together. We were trying to figure out how much money each part is worth when we divide \$100 into 5 equal parts. Can you show me how you would calculate that first step? \\
\bottomrule
\end{tabular}
\caption{Example benchmark dialogue showing real tutor performance evaluation. ``P'' is the action prompt variant, ``SF'' indicates if student followed their prompt (\cmark=yes), ``Rules'' shows which rules were checked, and ``Pass'' shows pass/fail for each rule (\cmark=pass, \xmark=fail). The tutor violated Rule 2 (multiple questions) and Rule 4 (ambiguity clarification) in Turn 2, and Rules 11-12 (accepting correct final answer) in Turn 10.}
\label{tab:benchmark_dialogue_example}
\end{table*}

\subsection{Analysis of Rule Violations}

This benchmark dialogue achieved an overall rule-following score of 86.2\% (25 out of 29 rule checks passed), demonstrating both strengths and weaknesses:

\textbf{Successful rule adherence (25 passes):}
\begin{itemize}
\item \textbf{Distraction handling (Turns 1, 5, 7):} The tutor consistently acknowledged distractions with witty replies and immediately redirected to the problem, following Rule 6 perfectly in all three instances.
\item \textbf{Socratic questioning (Turns 1-9):} Rule 1 was followed in 9 consecutive turns, with the tutor consistently avoiding revealing step elements prematurely while maintaining appropriate guidance.
\item \textbf{Emotional intelligence (Turn 6):} When the student expressed anxiety, the tutor provided empathetic acknowledgment before returning to problem-solving (Rule 7).
\item \textbf{Error handling (Turn 4):} The tutor correctly identified the student's misconception and provided constructive correction (Rule 9).
\end{itemize}

\textbf{Rule violations (4 failures):}
\begin{itemize}
\item \textbf{Turn 2 - Multiple questions (Rule 2 violation):} The tutor asked two questions in one sentence: ``What do you see in the problem about Gerald and Julia - what are they doing with the \$100?'' According to Rule 2, responses should end with only one leading question.
\item \textbf{Turn 2 - Missing clarification (Rule 4 violation):} When the student said ``I don't understand,'' the tutor should have asked for clarification (``What don't you understand?'') rather than guessing the student's meaning and proceeding with hints.
\item \textbf{Turn 10 - Rejecting correct answer (Rules 11-12 violation):} The student provided the correct final answer (\$50), but the tutor rejected it and asked to see intermediate calculations. Rule 11 requires accepting correct final answers without demanding step-by-step work, and Rule 12 requires concluding the problem with a solution summary.
\end{itemize}

\subsection{Comparison with Golden Dialogue}

Comparing this benchmark dialogue with the golden dialogue for the same problem (Appendix~\ref{app:golden_dialogue}) reveals interesting insights:

\begin{itemize}
\item Both dialogues contain the same sequence of triggered actions (7 distinct actions across 10 turns) since they follow a pregenerated ``master-plan'' of these sequences, which is determined based on a fixed random seed. This ensures consistent evaluation conditions across different tutors for the same problem.
\item The golden dialogue deliberately included anti-rule violations (Turns 2, 3, 6) for judge training, while the benchmark dialogue captured genuine tutor errors.
\item The benchmark tutor achieved 86.2\% rule adherence, showing that even state-of-the-art LLMs struggle with consistent pedagogical rule-following, particularly around ambiguity clarification and accepting final answers.
\item Both dialogues maintained pedagogical coherence despite systematic action triggering, validating our methodology's ability to create realistic evaluation scenarios.
\end{itemize}

\end{document}